\documentclass[letterpaper]{article} 
\usepackage{aaai23}  
\usepackage{times}  
\usepackage{helvet}  
\usepackage{courier}  
\usepackage[hyphens]{url}  
\usepackage{graphicx} 
\usepackage{natbib}  
\usepackage{caption} 
\usepackage{algorithm}
\usepackage{algorithmic}

\usepackage{amsfonts}
\usepackage{amsmath}
\usepackage{multirow}
\usepackage{tabularx}
\usepackage{makecell}
\usepackage{booktabs}
\usepackage{graphicx}

\usepackage{newfloat}
\usepackage{listings}
\DeclareCaptionStyle{ruled}{labelfont=normalfont,labelsep=colon,strut=off} 
\floatstyle{ruled}
\newfloat{listing}{tb}{lst}{}
\floatname{listing}{Listing}
\title{Learning to Memorize Entailment and Discourse Relations for Persona-Consistent Dialogues}
\author{
    Ruijun Chen\textsuperscript{\rm 1},
    Jin Wang\textsuperscript{\rm 1}\thanks{Corresponding author},
    Liang-Chih Yu\textsuperscript{\rm 2} and Xuejie Zhang\textsuperscript{\rm 1}
}
\affiliations{
    \textsuperscript{\rm 1}School of Information Science and Engineering, Yunnan University, Yunnan, China\\
    \textsuperscript{\rm 2}Department of Information Management, Yuan Ze University, Taiwan\\

    chenrj@mail.ynu.edu.cn, wangjin@ynu.edu.cn, lcyu@saturn.yzu.edu.tw, xjzhang@ynu.edu.cn
}

\usepackage{bibentry}

\begin{document}

\maketitle

\begin{abstract}
Maintaining engagement and consistency is particularly important in dialogue systems. Existing works have improved the performance of dialogue systems by intentionally learning interlocutor personas with sophisticated network structures. One issue with this approach is that it requires more personal corpora with annotations. Additionally, these models typically perform the next utterance prediction to generate a response but neglect the discourse coherence in the entire conversation. To address these issues, this study proposes a method of learning to memorize entailment and discourse relations for persona-consistent dialogue tasks. Entailment text pairs in natural language inference dataset were applied to learn latent entailment relations as external memories by premise-to-hypothesis generation task. Furthermore, an internal memory with a similar architecture was applied to the discourse information in the dialogue. Placing orthogonality restrictions on these two memory spaces ensures that the latent entailment relations remain dialogue-independent. Both memories collaborate to obtain entailment and discourse representation for the generation, allowing a deeper understanding of both consistency and coherence. Experiments on two large public datasets, PersonaChat and DSTC7-AVSD, demonstrated the effectiveness of the proposed method. Both automatic and human evaluations indicate that the proposed model outperforms several strong baselines in terms of both persona consistency and response coherence. Our source code is available at \url{https://github.com/Chenrj233/LMEDR}.
\end{abstract}

\begin{figure}[t]
\centering
\includegraphics[width=0.8\columnwidth]{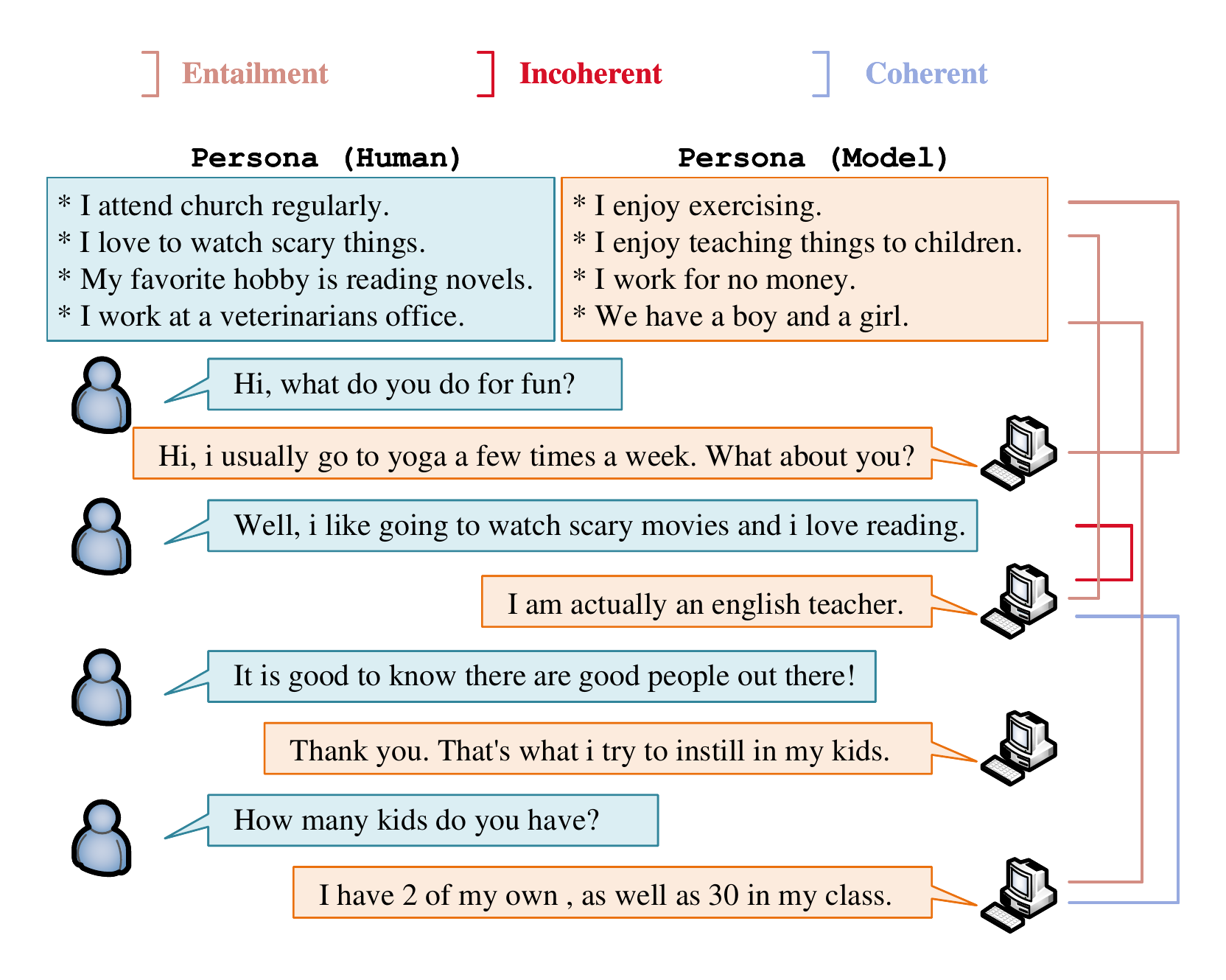} 
\caption{The conceptual diagram of introducing natural language inference in persona-based dialogue.}
\label{fig1}
\end{figure}
\begin{figure*}[t]
\centering
\includegraphics[width=0.6\textwidth]{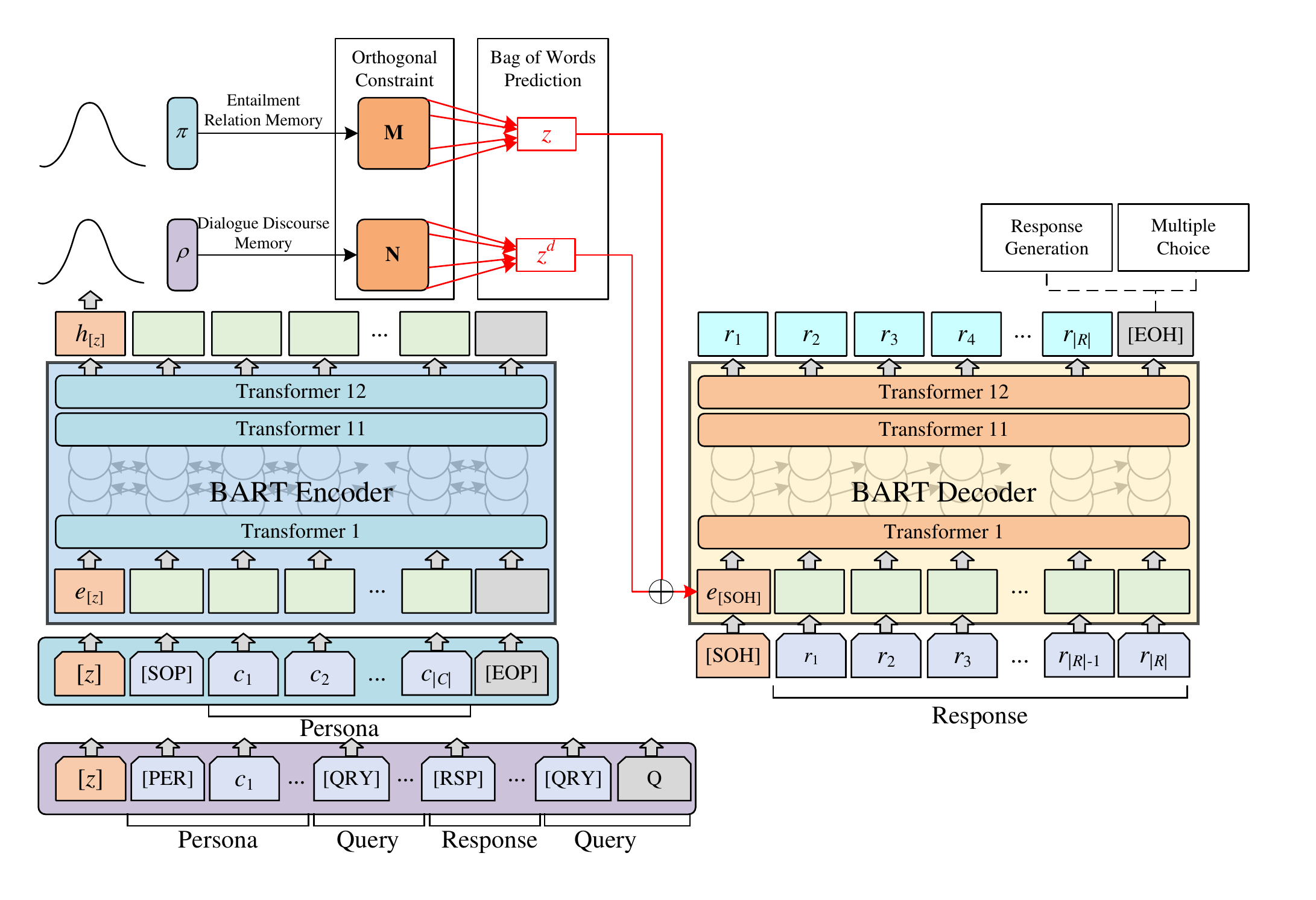} 
\caption{Overall architecture of the proposed method for persona-consistent dialogue generation.}
\label{fig2}
\end{figure*}

\section{Introduction}

Traditional chit-chat models lack specificity and personality consistency. Only when they access a sufficiently large dataset will they have the opportunity to generate piecemeal and uninformative responses in a chit-chat setting. For two consecutive questions with similar meanings in a two-round dialogue, that is, \textit{what is your job} and \textit{what do you do}, the model replied to the former: \textit{I am a lawyer}, while the latter: \textit{I am a doctor} \cite{Welleck2020}. This issue arises because of the lack of a consistent personality as well as an explicit memory towards plausibility as they are typically trained to produce a response given only recent dialogue history \cite{Shum2018}.

One solution that maintains consistency in a dialogue system is to provide a set of persona profiles that describe the character and then generate responses according to the persona. \textbf{Persona} can be defined as the composition of identity elements, such as profiles and background personal facts.

The expected outcome is that dialogue models generate a response consistent with the given persona. The PersonaChat dataset \cite{Zhang2018}, widely adopted to support the training of persona-consistent dialogues, was manually annotated by two annotators to act as part of a predefined persona and chat naturally to know each other during the conversation. However, given the time and effort needed to annotate more persona corpora to cover all possibilities, it is difficult to extend the application of such persona-related information to the daily usage of dialogue.

As humans, our knowledge of the concepts and the semantic relationship behind the language can allow us to rearrange unstructured data so that we can understand and analyze it. Essentially, we can robustly learn novel concepts with minimal supervision, benefitting from the well-known ability of natural language inference (NLI). Figure \ref{fig1} shows an example of introducing NLI in a persona-based dialogue. Given a persona as a \textit{premise}, we can determine whether the \textit{hypothesis} of the response utterance is true (entailment), false (contradiction), or undetermined (neutral). 

Recent studies have sought to improve the consistency of the dialogue system by modeling the understanding between interlocutors \cite{Liu2020}. \citet{Song2021} disentangled persona-based dialogue generation into two subtasks—response generation and consistency understanding—and used unlikelihood training to make the decoder generate contradictory dialogue responses as few as possible. However, multiple subtasks require multiple encoders, leading to a complex generation model structure. \citet{Nie2021_} introduced a contradiction detection task to evaluate the consistency in dialogues.

Despite continuing efforts to improve the engagement and consistency of dialogue systems, understanding persona-response consistency is still difficult. The key challenges are twofold: 1) Existing methods apply sophisticated structures to learn persona consistency, which requires more annotated corpora for training. However, persona-based corpora are still insufficient and difficult to collect. 2) Dialogue-generating models typically neglect discourse information. Discourse coherence is a crucial component of the effectiveness of a conversation, encompassing  how utterances are connected and how the entire dialogue is organized to convey information to the interlocutor. Existing models usually perform the next utterance prediction for response generation but ignore the dialogue discourse coherence. As indicated in Figure \ref{fig1}, \textit{I am actually an English teacher} seems to be an appropriate and persona-consistent response to the query. However, this response is incoherent in the context of an entire conversation.

To address these issues, this study proposes a method of learning to memorize entailment and discourse relations for persona-consistent dialogue tasks. We applied an encoder-decoder architecture from BART \cite{Lewis2020}. To explicitly understand the consistency of personas, we designed an external memory to store the latent entailment relations between premises and the entailment hypothesis, independent of dialogue itself. In addition, discourse relations were learned and stored in internal latent memory. The latent entailment relations are ensured to be dialogue-independent by imposing orthogonality constraints on the two memory spaces. Given personas and dialogue queries, both memories work jointly to obtain the entailment and discourse representation by the BART encoder. The generation  was finally accomplished by the BART decoder with two extra training objectives, which further acquired the ability to understand both consistency and coherence.

Comparative experiments were conducted using the PersonaChat \cite{Dinan2020} and DSTC7-AVSD \cite{Alamri2019}. Both automatic and human evaluations show that the proposed method generalizes well under different settings and outperforms several strong baselines on most metrics, especially persona consistency, indicating that the proposed method can produce better persona-consistent dialogue responses.

The remainder of this paper is organized as follows. Section 2 provides a brief review of the related work. Section 3 describes the proposed model, which learns to memorize entailment and discourse relations by using latent variables. Section 4 summarizes the specific experimental setup for the two public dialogue datasets and the corresponding analysis of the results. Finally, conclusions are drawn in Section 5.

\section{Related Work}
\subsection{Persona-based Dialogues}
Generation-based dialogue systems usually use the sequence-to-sequence (seq2seq) model \cite{Sutskever2014} as the backbone. After the persona is introduced into the dialogue, it is necessary to adopt an effective method to integrate role information into the dialogue, such as persona embedding \cite{Li2016}. Subsequently, with the development of large-scale pre-trained language models, an increasing number of methods \cite{Wolf2019,Roller2021,Lin2021,Zheng2020} have leveraged pre-training and fine-tuning to improve persona-based dialogue, but the problem of dialogue consistency remains unsolved. Therefore, \citet{Liu2020} have attempted to model the understanding between interlocutors to improve the consistency of dialogue systems. A new perspective \cite{Song2021} decomposes persona-based dialogue tasks into consistent understanding and dialogue generation significantly improves dialogue-consistent generation based on natural language inference.

\subsection{Latent Modeling}
In a dialogue scene, the factors that associate dialogue context with dialogue responses are often difficult to observe and explain; therefore, modeling the latent space of dialogue can help improve the performance of dialogue generation. Optimus \cite{Li2020} combines the advantages of BERT \cite{Devlin2019} and GPT-2 \cite{AlecRadfordJeffreyWuRewonChildDavidLuanDarioAmodei2020} for large-scale pre-training in the form of VAE \cite{Kingma2014} to model the latent variable space. PLATO \cite{Bao2020} introduces discrete latent variables to solve the one-to-many relationship in response generation. DialogVED \cite{Chen2022} introduces continuous latent variables into an enhanced encoder-decoder pre-training framework to improve the relevance and diversity of dialogue responses. All these methods show great promise for modeling dialogue-related features in latent space. This paper extents the idea by additionally memorizing NLI relations as latent dialogue-independent features.

\section{Learning to Memorize for Persona-consistent Dialogue}
The task of dialogue generation can be defined as the next utterance prediction, where a target response utterance ${\cal R} = [{r_1},{r_2},...,{r_{|{\cal R}|}}]$ is predicted given a conversation query ${\cal Q} = [{q_1},{q_2},...,{q_{|{\cal Q}|}}]$ according to given persona constraints ${\cal C} = [{c_1},{c_2},...{c_{{\rm{|}}{\cal C}{\rm{|}}}}]$. For convenience, the sentences $({\cal R},{\cal Q},{\cal C})$ are mapped to the vector representation $x=\{R,Q,C\}$. Further, natural language inference data \cite{Welleck2020,Williams2018} ${\cal N} = \{ {P^{(n)}},{H^{(n)}}\} _{n = 1}^N$, which consists of the entailed text pairs of premise and hypothesis, was used to learn the entailment relation to preserve consistency in dialogue generation.

Figure \ref{fig2} shows the overall architecture of the proposed learning to memorize the entailment and discourse relations model for a persona-consistent dialogue. The backbone model is based on BART \cite{Lewis2020}, which performs repeated two-stage  training, i.e., learning to memorize and persona-consistent dialogue generation. The key insight of the proposed model is that it maps both the entailment relation and discourse information to latent spaces. Based on this information, an external memory module enforces premise-to-hypothesis generation to map the textual entailed pair to the \textit{Dialogue-Independent} latent space, which can be memorized and stored in a memory structure ${\bf{M}}$. Similarly, the discourse information was mapped using an internal memory module ${\bf{N}}$ to learn the \textit{Dialogue-Related} features. For generation, both entailment and discourse representation can be obtained from memory and enhance persona consistency in dialogue generation with additional entailment and discourse information.

\begin{figure*}[t]
\centering
\includegraphics[width=0.6\textwidth]{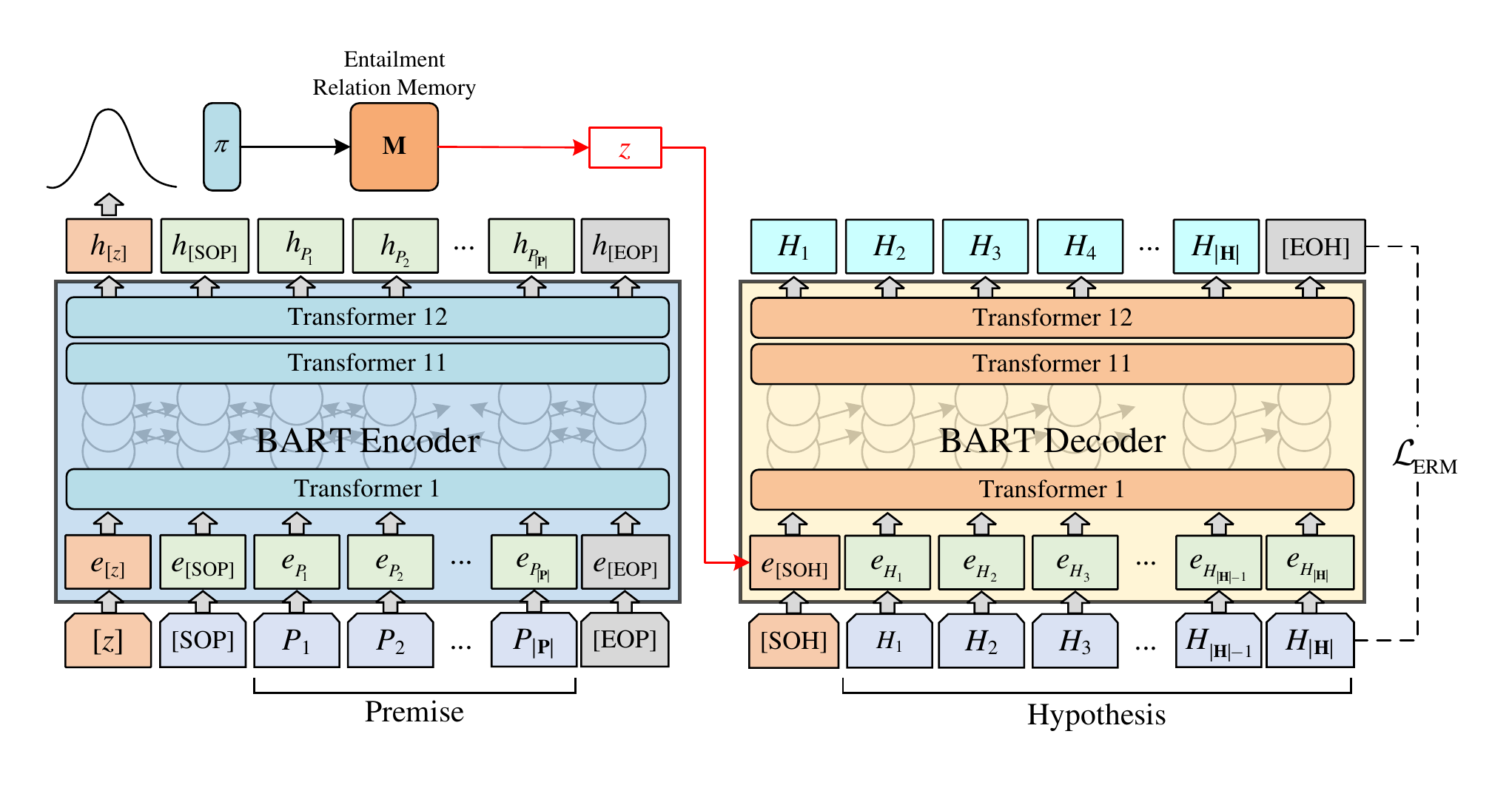} 
\caption{Learning to memorize the entailment relations in latent variables.}
\label{fig3}
\end{figure*}
\subsection{Learning to Memorize}
\subsubsection{Entailment Relation Memory (ERM).}
ERM is an external memory which is used to learn and store entailment relations for persona consistency. If a given hypothesis $H$ can be inferred from the premise $P$, the relationship of the pair is entailment. For persona-based dialogue, such an entailment relationship can be introduced to generate consistent responses. 

Given a dataset of textual entailed pairs ${\cal N} = \{ {P^{(n)}},{H^{(n)}}\} _{n = 1}^N$, textual entailment generation was adopted to learn a latent variable $z$, which represents the latent form of entailment relations in natural language inference, defined as
\begin{equation}
 p(H,z\left| P \right.) = p(z\left| P \right.)p(H\left| z \right., P{\rm{)}}
 \label{eq1}
 \end{equation}

 Based on the BART encoder, we introduce a special latent token $[z]$, a start-of-premise token $\rm{[SOP]}$, and an end-of-premise token $\rm{[EOP]}$ to the premise for latent entailment relation learning. By using the tokenizer and adding position embeddings, the input of the premise is transformed as
 \begin{equation}
 {{\bf{E}}_{{\rm{ERM}}}} = [{e_{[z]}},{e_{{\rm{[SOP]}}}},{e_{{p_1}}},{e_{{p_2}}},...,{e_{{p_{\left| P \right|}}}},{e_{{\rm{[EOP]}}}}]
 \label{eq2}
 \end{equation}

 We introduce a latent entailment relation memory structure ${\bf{M}}$ parameterized by $\theta$, as shown in Figure \ref{fig3}, where each element represents a certain latent factor, defined as
 \begin{equation}
{\mathbf{M}} = [{M_1},...,{M_k}] \in {\mathbb{R}^{k \times d}}
\label{eq3}
\end{equation}
where $k$ is the number of latent factors in entailment relations, and $d$ is the dimension of the memory element. The hidden state of the last layer of the BART encoder, that is, ${h_{[z]}}$ corresponding to ${e_{[z]}}$, was applied to learn the distribution of the latent entailment relations $z \sim p(z\left| P \right.)$ by
\begin{equation}
\pi  = softmax({W_\pi }{h_{[z]}} + {b_\pi })
\label{eq4}
\end{equation}
where $\pi$ represents the probability of each element in ${\bf{M}}$. Then, the latent entailment representation $z$ can be easily obtained from ${\bf{M}}$:
\begin{equation}
z = \sum\limits_{i = 1}^k {{\pi _i}{M_i}}
\label{eq5}
\end{equation}

To memorize the latent entailment relations, we use entailment representation $z$ from the memory ${\bf{M}}$ with the weights $\pi$, along with the premise to generate the corresponding hypothesis. The obtained entailment representation $z$ was added to the special start-of-hypothesis token $\rm{[SOH]}$ of the decoder, denote as
\begin{equation}
{\hat e_{{\rm{[SOH]}}}} = {e_{{\rm{[SOH]}}}} + z
\label{eq6}
\end{equation}

Latent memory can keep track of the entailment relation with the representation of the source premise by both reading and writing during generation. Notably, it can be updated by backpropagation of the premise-to-hypothesis generation.

The objective of the pre-training is to optimize memory ${\bf{M}}$ and model parameters $\varphi$ by minimizing the language modeling loss:
\begin{equation}
\begin{aligned}
  {\mathcal{L}_{ERM}} =  &  - {{\mathbb{E}}_{z \sim {p_{\theta ,\varphi }}(z\left| P) \right.}}\log {p_{\theta ,\varphi }}(H\left| {z,P)} \right. \\ 
   =  &  - {{\mathbb{E}}_{z \sim {p_{\theta ,\varphi }}(z\left| P \right.)}}\sum\limits_{t = 1}^{\left| H \right|} {\log {p_{\theta ,\varphi }}({H_t}\left| {z,P,{H_{ < t}})} \right.}  \\ 
\end{aligned}
\label{eq7}
\end{equation}

\begin{algorithm}[tb]
\caption{Latent memory learning}
\label{alg:algorithm1}
\textbf{Input}: A set of entailed text pair ${\cal N} = \{ {P^{(n)}},{H^{(n)}}\} _{n = 1}^N$ and persona dialogue dataset ${\bf{x}} = \{ R,Q,C\} $.\\
\textbf{Parameter}: Memory ${\bf{M}}$, ${\bf{N}}$ parameterized by $\theta$, $\phi$, a pre-trained BART parameterized by $\varphi$.
\begin{algorithmic}[1] 
\REPEAT
\STATE \textbf{Stage 1}:
\STATE Train the BART model and memory ${\bf{M}}$ with input  ${\cal N}$.
\STATE Minimizing Eq. (\ref{eq7}) and optimize $\theta$, $\varphi$.
\STATE \textbf{Stage 2}:
\STATE Fixed parameter $\theta $.
\STATE Train the BART model and memory ${\bf{N}}$ with input $x$.
\STATE Minimizing Eq. (\ref{eq19}) and optimize  $\phi$, $\varphi$.
\UNTIL convergence
\end{algorithmic}
\end{algorithm}

\subsubsection{Dialogue Discourse Memory (DDM).}
Discourse coherence is an important aspect of dialogue text quality. It encompasses how utterances are connected, as well as how the entire dialogue is organized to convey information to the interlocutor \cite{Bao2020}. Similarly, we memorized the discourse information into an internal memory ${\bf{N}}$ parameterized by $\phi$. Let ${\cal C} = [{c_1},{c_2},...{c_{{\rm{|}}{\cal C}{\rm{|}}}}]$ denote the persona of the agent, ${{\cal Q}^{(1)}},...,{{\cal Q}^{(m)}}$ represent $m$ dialogue queries, and ${{\cal R}^{(1)}},...,{{\cal R}^{(m)}}$ represent the target responses. Here, ${\cal Q}$ and ${\cal R}$ are consecutive context-response pairs from the same dialogue session, which are used to capture the correct discourse relation in a dialogue. The input of the BART encoder is the concatenation of the persona and dialogue content, denoted as 
\begin{equation}
\begin{array}{l}
{{\bf{E}}_{{\rm{DDM}}}} = [{e_{[z]}},{e_{{\rm{[PER]}}}},{\cal C},{e_{{\rm{[QRY]}}}},{{\cal Q}^{(1)}},{e_{{\rm{[RSP]}}}},{{\cal R}^{(1)}},\\
\;\;\;\;\;\;\;\;\;\;\;\;\;\;\;\;\;\;\;\;\;\;\;\;\;\;\;\;\;\;\;\;\;\;\;\;\;\;\;\;\;\;...,{e_{{\rm{[QRY]}}}},{{\cal Q}^{(m)}}]
\end{array}
\label{eq8}
\end{equation}
where \rm{[QRY]} and \rm{[RSP]} are two special tokens that indicate the beginning of the query and the response, respectively.

Latent memory ${\mathbf{N}} \in {\mathbb{R}^{l \times d}}$ was introduced, where $l$ represents the kind of latent dialogue discourse information. It learns the distribution of the latent dialogue discourse using
\begin{equation}
\rho  = softmax({W_\rho }{h_{[z]}} + {b_\rho })
\label{eq9}
\end{equation}

Then, the latent dialogue discourse representation was calculated by
\begin{equation}
{z^d} = \sum\limits_{j = 1}^l {{\rho _j}{N_j}}
\label{eq10}
\end{equation}
\begin{table*}[t]
  \centering
    \begin{tabular}{lllllll}
    \toprule
    \multirow{2}[4]{*}{Model} & \multicolumn{3}{p{12.57em}}{Original} & \multicolumn{3}{p{12.57em}}{Revised} \\
\cmidrule{2-7}    \multicolumn{1}{c}{} & \multicolumn{1}{p{4.19em}}{Hits@1} & \multicolumn{1}{p{4.19em}}{PPL} & \multicolumn{1}{p{4.19em}}{F1} & \multicolumn{1}{p{4.19em}}{Hits@1} & \multicolumn{1}{p{4.19em}}{PPL} & \multicolumn{1}{p{4.19em}}{F1} \\
    \midrule
    KV Profile Memory & 54.8  &  -    & 14.25 & 38.1  &  -    & 13.65 \\
    Generative Profile Memory & 10.2  & 35.01 & 16.29 & 9.9   & 34.94 & 15.71 \\
    LIC   & 17.3  &   -   & 17.79 & 16.2  &   -   & 16.83 \\
    Transfertransfo & 82.1  & 17.51 & 19.09 &   -   &   -   &  - \\
    P$^2$BOT & 81.9  & 15.12 & 19.77 & 68.6  & 18.89 & 19.08 \\
    Our   & \textbf{89.5} & \textbf{10.99} & \textbf{21.99} & \textbf{85.0} & \textbf{13.42} & \textbf{19.54} \\
    \bottomrule
    \end{tabular}
\caption{Automatic evaluation results of different methods for persona-based dialogue generation.}
\label{table1}
\end{table*}

In addition, latent entailment relations should be independent of the dialogue context. Therefore, we propose imposing an orthogonal constraint on these two memory spaces to reduce the correlation between the different latent spaces. The orthogonal constraint can encourage the latent memory to learn more features and reduce redundant features. To accomplish this goal, we used cosine similarity to represent the relationship between them, described as
\begin{equation}
\cos ({\bf{M}},{\bf{N}}) = \frac{{{\bf{M}}{{\bf{N}}^ \top }}}{{{{\left\| {\bf{M}} \right\|}_2}{{\left\| {\bf{N}} \right\|}_2}}}
\label{eq11}
\end{equation}
where ${\left\|  \cdot  \right\|_2}$ represents ${L_2}$ normalization. Once the vectors of different latent-memory spaces are orthogonal to each other, the correlation between them is the lowest. Thus, the training objective of the DDM can be defined to minimize the cosine similarity between ${\bf{M}}$ and ${\bf{N}}$: 
\begin{equation}
{{\cal L}_{{\rm{DDM}}}} = {\sum\limits_{i \le k,j \le l} {\left( {\frac{{{M_i}N_j^ \top }}{{{{\left\| {{M_i}} \right\|}_2}{{\left\| {{N_j}} \right\|}_2}}}} \right)} ^2}
\label{eq12}
\end{equation}

\subsection{Persona-consistent Responses Generation}
For generation, the model leverages latent memories of both entailment relations and dialogue discourse to preserve consistency for generating a persona-based response, denoted as
\begin{equation}
p\left( {R,z,{z^d}\left| {Q,C} \right.} \right) = p(R\left| {z,{z^d},Q,C)p(z\left| {C)p({z^d}\left| {Q,C)} \right.} \right.} \right.
\label{eq13}
\end{equation}

We applied the persona as a premise and fed it into the model and obtained the latent entailment relations representation $z$ under the condition of persona from latent entailment memory ${\bf{M}}$. The latent entailment relation representation $z$ and the latent dialogue discourse representation ${z^d}$ are then added to the special start-of-hypothesis token ${e_{{\rm{[SOH]}}}}$ of the decoder input, denoted as
\begin{equation}
{\hat e_{{\rm{[SOH]}}}} = {e_{{\rm{[SOH]}}}} + z + {z^d}
\label{eq14}
\end{equation}

Additionally, we use a bag-of-words loss \cite{Zhao2017} to facilitate the latent variable,
\begin{equation}
\resizebox{1\hsize}{!}{$
\begin{aligned}
{\mathcal{L}_{{\text{BOW}}}} = &- {\mathbb{E}_{z \sim {p_\varphi }(z\left| {C),{z^d} \sim {p_{\phi ,\varphi }}(} \right.{z^d}\left| {C,Q)} \right.}} \sum\limits_{t = 1}^{\left| R \right|} {\log {p_{\phi ,\varphi }}} ({R_t}\left| {C,Q,z,{z^d})} \right. \\ 
   = &  - {\mathbb{E}_{z \sim {p_\varphi }(z\left| {C),{z^d} \sim {p_{\phi ,\varphi }}(} \right.{z^d}\left| {C,Q)} \right.}}\sum\limits_{t = 1}^{\left| R \right|} {\log \frac{{{e^{f({R_t})}}}}{{\sum\limits_{v \in V} {{e^{f\left( v \right)}}} }}} \\ 
\end{aligned}
$}
\label{eq15}
\end{equation}
where $V$ is the whole vocabulary and $f\left( v \right)$ represents the predicted probability of token $v$  appearing in the target response.

For response generation, we also trained the parameters of the model using a language modeling loss function:
\begin{equation}
\resizebox{1\hsize}{!}{$
\begin{aligned}
  {\mathcal{L}_{{\text{LM}}}} = & - {\mathbb{E}_{z \sim {p_\varphi }(z\left| {C),{z^d} \sim {p_{\phi ,\varphi }}(} \right.{z^d}\left| {C,Q)} \right.}}\log {p_{\phi ,\varphi }}(R\left| {C,Q,z,{z^d}} \right.) \\ 
 = & - {\mathbb{E}_{z \sim {p_\varphi }(z\left| {C),{z^d} \sim {p_{\phi ,\varphi }}(} \right.{z^d}\left| {C,Q)} \right.}}\sum\limits_{t = 1}^{\left| R \right|} {\log {p_{\phi ,\varphi }}({R_t}\left| {{R_{ < t}},C,Q,z,{z^d}})\right.}  \\ 
\end{aligned}
$}
\label{eq16}
\end{equation}

Similar to \cite{Wolf2019}, we randomly sampled $t$ interference responses to train the model to select the correct response, i.e., as a multiple-choice task. Specifically, we use the hidden state of the last token output by the decoder to predict the scores ${\hat y_i}$ of each candidate response and calculate the cross-entropy loss with the ground-truth label ${y_i}$, denoted as
\begin{equation}
\hat y = {\mathop{\rm softmax}\nolimits} ({W_h}{h_{eos}} + {b_h})
\label{eq17}
\end{equation}
\begin{equation}
{{\cal L}_{{\rm{CLS}}}} =  - \sum\limits_{i = 1}^{t+1} {{{\hat y}_i}\log ({y_i})}
\label{eq18}
\end{equation}

In summary, the objective of our model is to minimize the loss:
\begin{equation}
{\cal L}(\phi ,\varphi ) = {{\cal L}_{{\rm{DDM}}}} + {{\cal L}_{{\rm{BOW}}}} + {{\cal L}_{{\rm{LM}}}} + {{\cal L}_{{\rm{CLS}}}}
\label{eq19}
\end{equation}

While training persona-consistent responses generation, we fixed the parameters of ${\bf{M}}$; the specific process is described in Algorithm \ref{alg:algorithm1}.

\begin{table}[t]
  \centering
    \begin{tabular}{llll}
    \toprule
    Model & Hits@1 & PPL   & F1 \\
    \midrule
    PE-Trans & \multicolumn{1}{l}{89.4} & \multicolumn{1}{l}{-} & \multicolumn{1}{l}{-} \\
    ImageS2S & \multicolumn{1}{l}{-} & \multicolumn{1}{l}{11.19} & \multicolumn{1}{l}{21.30} \\
    BART  & \multicolumn{1}{l}{86.9} & \multicolumn{1}{l}{11.85} & \multicolumn{1}{l}{20.72} \\
    Our   & 89.5(\textbf{90.1}) & \textbf{10.99}(11.00) & \textbf{21.99}(21.96) \\
    \bottomrule
    \end{tabular}%
\caption{Comparison with pre-trained language models fine-tuned on PersonaChat original mode.}
\label{table2}%
\end{table}%
\begin{table}[t]
  \centering
    \begin{tabular}{llll}
    \toprule
    Model & \multicolumn{1}{p{4.19em}}{Dist-1} & \multicolumn{1}{p{4.19em}}{Dist-2} & \multicolumn{1}{p{4.19em}}{C.Score} \\
    \midrule
    LIC   & 2.31  & 10.71 & 19.13 \\
    P$^2$BOT & 1.87   & 10.08   &  23.84\\
    BOB   & \textbf{2.59}      & \textbf{13.90}   &  22.45\\
    Our   & 2.47  & 13.82 &  \textbf{25.31}\\
    \bottomrule
    \end{tabular}
\caption{Automatic evaluation results of different methods for persona consistency.}
\label{table3}
\end{table}
\begin{table*}[t]
  \centering
    \begin{tabular}{llllllll}
    \toprule
    Model & \multicolumn{1}{p{4.19em}}{BLEU-1} & \multicolumn{1}{p{4.19em}}{BLEU-2} & \multicolumn{1}{p{4.19em}}{BLEU-3} & \multicolumn{1}{p{4.19em}}{BLEU-4} & \multicolumn{1}{p{4.19em}}{METEOR} & \multicolumn{1}{p{4.49em}}{ROUGE-L} & \multicolumn{1}{p{4.19em}}{CIDEr} \\
    \midrule
    CMU   & 0.718 & 0.584 & 0.478 & 0.394 & 0.267 & 0.563 & 1.094 \\
    PLATO & 0.784 & 0.637 & 0.525 & 0.435 & 0.286 & 0.596 & 1.209 \\
    ProphetNet & \textbf{0.824} & 0.691 & \textbf{0.582} & 0.487 & 0.313 & 0.635 & 1.382 \\
    DialogVED & 0.822 & \textbf{0.692} & \textbf{0.582} & \textbf{0.489} & 0.312 & \textbf{0.636} & 1.391 \\
    Our   & 0.801 & 0.680  & 0.576 & 0.488 & \textbf{0.316} & 0.631 & \textbf{1.403} \\
    \bottomrule
    \end{tabular}
\caption{Automatic evaluation results of different methods for DSTC7-AVSD.}
\label{table4}
\end{table*}
\begin{table}[t]
  \centering
    \begin{tabular}{llll}
    \toprule
    Model & \multicolumn{1}{p{4.19em}}{Fluency} & \multicolumn{1}{p{4.19em}}{Consistency} & \multicolumn{1}{p{4.19em}}{Avg} \\
    \midrule
    LIC   &  3.27  &  2.11 & 2.69 \\
    P$^2$BOT &  3.51  &  2.20 &  2.86\\
    Our   &  \textbf{3.57} &   \textbf{2.31} &  \textbf{2.94}\\
    \bottomrule
    \end{tabular}%
\caption{Human evaluation results.}
\label{table5}%
\end{table}%
\begin{table}[t]
  \centering
    \begin{tabular}{lllll}
    \toprule
    \multicolumn{1}{l}{} & \multicolumn{1}{l}{Hits@1} & \multicolumn{1}{l}{F1} & \multicolumn{1}{l}{BLEU-4} & \multicolumn{1}{l}{C.Score} \\
    \midrule
    BART  & 86.9  & 20.72 & 0.01289 &  21.32\\
    w/o ERM & 89.3  & 21.70  & 0.01597 & 22.09 \\
    w/o DDM & 88.4  & 21.72 & 0.01406 &  24.66\\
    w/o OC & 88.8  & 21.84 & 0.01485 &  25.01\\
    Our   & 89.5  & 21.99 & 0.01561 &  25.31\\
    \bottomrule
    \end{tabular}%
\caption{Analysis of ablation experiments on the PersonaChat original mode.}
\label{table6}%
\end{table}%
\begin{table*}[t]
  \centering
    \begin{tabular}{llllllll}
    \toprule
    Model & \multicolumn{1}{p{4.19em}}{BLEU-1} & \multicolumn{1}{p{4.19em}}{BLEU-2} & \multicolumn{1}{p{4.19em}}{BLEU-3} & \multicolumn{1}{p{4.19em}}{BLEU-4} & \multicolumn{1}{p{4.19em}}{METEOR} & \multicolumn{1}{p{4.49em}}{ROUGE-L} & \multicolumn{1}{p{4.19em}}{CIDEr} \\
    \midrule
    BART  &  0.791 &  0.668 &  0.562 & 0.473  & 0.307  & 0.619  & 1.372 \\
    w/o ERM & 0.795  & 0.671  & 0.565  & 0.477  &  0.313 & 0.625  & 1.395 \\
    w/o DDM & 0.799  &  0.676 & 0.571  & 0.482  &  0.309 &  0.625 & 1.388 \\
    w/o OC & 0.782 & 0.660 & 0.557 & 0.473 & 0.309 & 0.620 & 1.388 \\
    Our   & 0.801 & 0.680  & 0.576 & 0.488 & 0.316 & 0.631 & 1.403 \\
    \bottomrule
    \end{tabular}
\caption{Analysis of ablation experiments on the DSTC7-AVSD.}   
\label{table7}
\end{table*}

\section{Experiments}
\subsection{Dataset}
\subsubsection{Dialogue Dataset.}
We conducted experiments on two publicly available dialogue datasets to evaluate the performance of the proposed method. 
\begin{itemize}
    \item \textbf{ConvAI2 PersonaChat} \cite{Dinan2020} is a chitosan–chat dataset based on PersonaChat \cite{Zhang2018}. It provides the personas of the interlocutor and is designed to facilitate dialogue models to generate more consistent and engaging responses.
    \item \textbf{DSTC7-AVSD} \cite{Alamri2019} provides a conversational question-answering dataset, which is similar to that of PersonaChat. We used this dataset to explore the performance of our proposed method on a contextual knowledge-based dialogue task. The system must generate answers based on the given context and background knowledge of the conversation. We used only text information for the experiments.
\end{itemize}
\subsubsection{NLI Dataset.}
Additionally, two natural language inference datasets were used to learn the corresponding ERM.
\begin{itemize}
    \item \textbf{DNLI} \cite{Welleck2020} is a dialogue inference dataset based on PersonaChat. The dataset consisted of sentence pairs labeled as entailment, neutral, or contradictory.
    \item \textbf{MNLI} \cite{Williams2018} is a multi-genre natural language inference corpus and one of the largest corpora available for recognizing textual entailments.
\end{itemize}

\subsection{Compared Methods}
\subsubsection{Baseline.}
Both \textbf{KV profile memory} and \textbf{generative profile memory} \cite{Zhang2018} are the official baselines for PersonaChat, whereas \textbf{LIC} \cite{Golovanov2020} was the best performing model in ConvAI2 for human evaluation, and \textbf{Transfertransfo} \cite{Wolf2019} achieved the best performance on automatic metrics in ConvAI2. For DSTC7-AVSD, \textbf{CMU} \cite{Sanabria2018} achieved the best performance for all evaluation metrics in the competition.
\subsubsection{State of the art.}
\textbf{P$^2$BOT} \cite{Liu2020} models the understanding between the interlocutors and obtains a new state-of-the-art on ConvAI2. \textbf{Poly-Encoder Transformer} (PE-Trans) \cite{Humeau2019} was pre-trained on Reddit and fine-tuned on ConvAI2 to obtain the best performance in the response selection task. \textbf{Imageseq2Seq Dodecadi-alogue} (ImageS2S) \cite{Shuster2020} was trained on 12 tasks and fine-tuned on the ConvAI2 generation task. \textbf{BOB} \cite{Song2021} disentangles persona-based dialogue generation into consistent understanding and dialogue generation tasks to ensure that the model has a better consistent understanding. For DSTC7-AVSD, \textbf{PLATO} \cite{Bao2020} was the first large-scale pre-trained dialogue language model that introduced a discrete variable for one-to-many relationship modeling. \textbf{ProphetNets} \cite{Qi2020} proposed a pre-training objective for predicting multiple future tokens to enhance the performance of pre-trained language models on natural language generation tasks. \textbf{DialogVED} \cite{Chen2022} is a pre-trained model framework to enhance the encoder-decoder by introducing continuous latent variables and obtaining the state-of-the-art on the DSTC7-AVSD.

\subsection{Implementation Details}
The proposed model was initialized using BART-large. AdamW \cite{Loshchilov2017} was applied to optimize the model, with an initial learning rate of 8e-6. We used \textbf{DNLI} on \textbf{PersonaChat} and \textbf{MNLI} on \textbf{DSTC7-AVSD} for the ERM learning. The batch size was 64 for training stage 1, and we used a batch size of two with a gradient accumulation of eight for training stage 2. The types of ERM and DDM were both set to 10 for PersonaChat and set to 20 and 5 for DSTC7-AVSD. The proposed model trained on one NVIDIA RTX 3090 with PyTorch framework. For dialogue generation, we used a beam search, and the maximum sequence length was set to 50. 

\subsection{Automatic Evaluation}
Following the official automatic evaluation, we used \textbf{Hits@1}, \textbf{Perplexity} (PPL), and \textbf{F1} for automatic evaluation on the \textbf{PersonaChat} dataset. \textbf{Hits@1} is the probability that the golden response ranks the highest among the 20 candidate responses. \textbf{Perplexity} was used to calculate the negative log-likelihood of the golden response from the model. \textbf{F1} is calculated from the precision and recall at the word level between the predicted  and golden responses. For persona consistency, we apply \textbf{Consistency Score} (C.Score) \cite{Madotto2020}, which leverages a referee model to predict consistency between response and persona. \textbf{Dist-1/2} \cite{Li2016a} is used to measure response diversity.

The \textbf{BLEU}, \textbf{METEOR}, \textbf{ROUGE-L}, and \textbf{CIDEr} metrics were reported for \textbf{DSTC7-AVSD} automatic evaluation, similar to DSTC7 reviews \cite{Alamri2019}. 

Table \ref{table1} presents the automatic evaluation results of the different methods for persona-based dialogue generation. As indicated, the proposed model outperformed all the baselines on the PersonaChat dataset, especially on the revised dataset. The proposed model achieved significant improvements in response selection tasks because ERM can provide more entailed information about the persona while DDM can captures the appropriate discourse information in the dialogue, making it easier to distinguish the correct response from the candidate responses. The improvement in PPL and F1 also shows that ERM and DDM can further improve the consistency of the persona and the quality of the response.

Table \ref{table2} compares the results of the proposed method with some pre-trained language models fine-tuned on the PersonaChat. The results of training with 14 interference responses are in parentheses. The proposed model also achieves better results compared to models trained on larger corpora and more tasks. Additionally, adding more interference responses can improve the accuracy of the model to select the correct response. Table \ref{table3} shows the result of persona consistency. The proposed model still obtained the highest C.Score, which indicates that the responses generated by the proposed model perform the best on persona consistency.

Table \ref{table4} shows the experimental results of DSTC7-AVSD. The proposed model achieves the best results for METROR and CIDEr and is close to the best model for the other metrics. Better performance on CIDEr shows that in background-based dialogue question answering, the proposed model can use ERM to capture key information in the background and combine the dialogue history to generate high-quality responses. 

\subsection{Human Evaluation}
We conducted a human evaluation of the state-of-the-art models (LIC, P$^2$BOT) and the proposed method. We randomly sampled 100 responses generated by these models from the original PersonaChat dev set for the human evaluation. Both \textbf{fluency} and \textbf{consistency} were applied as criteria. Four human annotators were asked to rate \textbf{fluency} on a scale of 1 to 5 and \textbf{consistency} on a scale of 1 to 3, where the Fleiss’s kappa of fluency and consistency are 0.578 and 0.671. Here, \textbf{fluency} indicates the smoothness of responses and conversation, coherence is included, where 1 means terrible and 5 represents very satisfying. \textbf{Consistency} represents the consistency between persona and response, which reflects whether the model can maintain persona consistency, where 1 means it does not  match persona, 2 means it is irrelevant, and 3 means it is consistent with persona. As shown in Table \ref{table5}, the results of the human evaluation are consistent with the automatic evaluation, and our model outperforms the previous best-performing model on human evaluation in terms of both fluency and consistency. Several examples of the generated responses are provided in the Appendix to help illustrate the effectiveness of our model. 

\subsection{Ablation Study}
We conducted ablation experiments on the PersonaChat and DSTC7-AVSD to explore the impact of each module. The ablation results are presented in Tables \ref{table6} and \ref{table7}, respectively. 

\subsubsection{Effect of ERM.}
After removing the ERM, the C.Score becomes lower on PersonaChat, which shows that the ERM can make the response generated by the model more consistent with its persona and improve the persona consistency in the dialogue response. In dialogue question answering, ERM can capture the key information in the background so that the generated answers can achieve better results for CIDEr.

\subsubsection{Effect of DDM.}
The large decline in BLEU-4 indicates that the quality of the responses generated by the model deteriorates without DDM. The metric of METEOR has a high correlation with the results of human judgment. Without DDM, METEOR drops to a large extent, validating the important role that DDM plays in response generation and DDM can effectively capture the connections of the utterances in the dialogue, thereby making the responses more coherent and natural.

\subsubsection{Effect of the Orthogonal Constraint (OC).}
Imposing orthogonality constraints on the two latent spaces effectively reduces redundant features and makes the features captured by the model easier to distinguish, thereby improving the quality of generation and performance in response selection.

\section{Conclusion}
In this paper, we propose a dialogue generation method for learning to memorize entailment and discourse relations with latent variables. Combining latent entailment relations and dialogue discourse relations makes generated responses more coherent and consistent. Experiments on the PersonaChat dataset demonstrate the effectiveness of the proposed method. The results on the DSTC7-AVSD dataset also show that learning entailment and discourse relations are beneficial for dialogue question-answering generation. 

Future works will attempt to explore different latent relations in text pairs on different datasets and combine the discourse relations in the dialogue to make the dialogue generation in the desired direction.

\section*{Acknowledgments}
This work was supported by the National Natural Science Foundation of China (NSFC) under Grant Nos. 61966038 and 62266051, the Ministry of Science and Technology, Taiwan, ROC, under Grant No. MOST 111-2628-E-155-001-MY2 and the Postgraduate Research and Innovation Foundation of Yunnan University under Grant No.2021Z076. The authors would like to thank the anonymous reviewers for their constructive comments.

\bibliography{ref}
\appendix
\section{Appendix}
\subsection{Case Study}
\begin{table*}[ht]
  \centering
    \begin{tabular}{ll}
    \toprule
    \multirow{4}{*}{Persona} & i listen to rap music. \\
    ~& i produce music for artists. \\
    ~& i drive a 2015 honda civic. \\
    ~& my favourite food is pizza. \\
    \midrule
    \multirow{6}{*}{Context} & Q: hi , how are you ? do you have any brothers or sisters ? \\
    ~& R: no i don't do you ? \\
    ~& Q: yes , i'm 13 and i've an older brother . \\
    ~& R: that's nice what kind of music do you like \\
    ~& Q: i do not have much time as i play soccer . you ? \\
    ~& R: i am a music producer for rap artists \\
   \bottomrule
\toprule    Query & \textbf{cool i like rap . i hate maths though ! do you have other hobbies} \\
\midrule   GOLD  & work takes up a lot of my time \\
\midrule  LIC   & i love to eat pizza . \\
\midrule   BoB   &  i like music and i like to listen to music (\textbf{Incoherent with query})\\
\midrule    P$^2$BOT & i like to listen to rap music (\textbf{Incoherent with query})\\
\midrule    Our   & i like to drive my honda civic \\
\bottomrule    \end{tabular}%
\caption{Case analysis of response generation on the PersonaChat.}
\label{table8}%
\end{table*}%

\begin{table*}[ht]
  \centering
    \begin{tabular}{ll}
    \toprule
    \multirow{5}{*}{Persona} & my dream car is a rolls royce ghost. \\
    ~ & i currently work for a watch dealer.\\
    ~ & my family does not support my career choices. \\
    ~ & i often shop for shoes and watches. \\
    ~ & i like shopping. \\
    \midrule
    \multirow{4}{*}{Context} & Q: good evening . what are you up to ? \\
    ~ & R: hello , how are you ? i am well . i like to shop for shoes and watches . \\
    ~ & Q: i don't shop much . prefer to relax at home with my cow . \\
    ~ & R: what do you do ? i'm work for a watch dealer . \\
   \bottomrule
    \toprule
    Query & \textbf{no wonder you like watches ! i'm a commercial electrician} \\
\midrule   GOLD  & wonderful ! yeah , however , my family does not support my choice of career .\\
\midrule   LIC   & i love watching movies (\textbf{Irrelevant to persona})\\
\midrule    BoB   &  i work at a dealership, but i am not very good at it.\\
\midrule   P$^2$BOT & what do you do for a living ? i work for the watch . (\textbf{Incoherent with context})\\
\midrule    Our   & my family does not support my career choices . \\
\bottomrule
\toprule    Query & \textbf{why not ? family is important .} \\
\midrule   GOLD  & i'm not sure why they do not . do you like cars ? \\
\midrule  LIC   & i like to watch movies (\textbf{Irrelevant to persona})\\
\midrule   BoB   &  i am not sure. i am a watch dealer. (\textbf{Incoherent with context})\\
\midrule    P$^2$BOT & my family doesn ' t support my lifestyle choices . (\textbf{Incoherent with context})\\
\midrule    Our   & i do not know . i do have a dream car though . it is a rolls royce ghost .\\
\bottomrule    \end{tabular}%
\caption{Case analysis of response generation on the PersonaChat.}
\label{table9}%
\end{table*}%

\begin{table*}[ht]
  \centering
    \begin{tabular}{lp{30em}}
    \toprule
    Caption & a man closes his window , then he sneezes twice before taking a drink . then he opens up a bag and digs through it looking for something before walking out of the room . \\
    \midrule
    Summary & a man closes the window , goes to the table and goes through the items in a bag , takes a drink from the green cup and leaves the room . \\
    \midrule
    \multirow{3}[6]{*}{Context} & Q: what is the guy doing at the window? \\
\cmidrule{2-2}     & R: the guy is closing the window \\
\cmidrule{2-2}     & Q: what does he do after that? \\
    \bottomrule
    \toprule
    Baseline & he picks up a book from the table \\
    \midrule
    PLATO & he goes to the table and takes a drink from a green cup \\
    \midrule
    DialogVED & he goes to the table and goes through the items in a bag before taking a drink \\
    \midrule
    Our   & he sneezes twice and then takes a drink \\
    \bottomrule
    \end{tabular}%
\caption{Case analysis of response generation on the DSTC7-AVSD.}
\label{table10}%
\end{table*}%
Tables \ref{table8} and \ref{table9} present the responses generated by the different methods on PersonaChat. As indicated, the responses generated by the proposed method were more consistent with their personas and coherence on PersonaChat. In Table \ref{table8}, the responses generated by P$^2$BOT and BoB are inconsistent with \textit{other hobbies} in the query, even though the responses are consistent with the persona. Both LIC and our model generate more coherent responses based on other persona. As shown in Table \ref{table9}, the responses generated by LIC are irrelevant to the personas. P$^2$BOT and BoB tend to ignore content from dialogue, generating repetitive responses that make the entire conversation incoherent. Our model can effectively combine dialogue content and persona to generate appropriate responses.

Table \ref{table10} provides the responses generated on DSTC7-AVSD, where the models need to generate responses based on the given background knowledge and dialogue content. It can be found that our model is able to combine dialogue content to generate answers with key information in the context that is closely related to the query.

\end{document}